\documentclass[conference]{IEEEtran}
\IEEEoverridecommandlockouts

\usepackage{textcomp}
\usepackage{amssymb}
\usepackage{mathtools}
\usepackage{mathrsfs}
\usepackage{graphicx}
\usepackage[space]{grffile}
\usepackage[utf8]{inputenc}
\usepackage[T1]{fontenc}
\usepackage{url}
\usepackage{booktabs}
\usepackage{amsfonts}
\usepackage{nicefrac}
\usepackage[table]{xcolor}
\usepackage{wrapfig}
\usepackage[numbers,sort&compress]{natbib}
\usepackage{subfigure}
\usepackage{caption}
\usepackage{enumitem}
\usepackage{amsmath}
\usepackage{tikz-cd}
\usepackage[colorlinks=true,citecolor=brown,linkcolor=blue,urlcolor=blue]{hyperref}
\usepackage{colortbl}
\usepackage{algorithm}
\usepackage{algpseudocode}
\usepackage{bbding}
\usepackage{fancyvrb}
\usepackage{listings}
\usepackage{diagbox}
\usepackage{multirow}

\usepackage{tcolorbox}

\newtcolorbox{promptbox}[1][]{
    colback=red!3,          %
    colframe=red!40,        %
    arc=3mm,
    title=#1,
    fonttitle=\bfseries,
    boxrule=1pt,
    enhanced
}

\newtcolorbox{responsebox}[1][]{
    colback=blue!3,         %
    colframe=blue!40,       %
    arc=3mm,
    title=#1,
    fonttitle=\bfseries,
    boxrule=1pt,
    enhanced
}

\usepackage[textsize=tiny,disable]{todonotes}

\def\BibTeX{{\rm B\kern-.05em{\sc i\kern-.025em b}\kern-.08em
    T\kern-.1667em\lower.7ex\hbox{E}\kern-.125emX}}

\definecolor{codegreen}{rgb}{0,0.5,0}
\definecolor{codered}{rgb}{0.7,0.1,0.1}
\definecolor{codepurple}{rgb}{0.58,0,0.82}
\definecolor{backcolour}{rgb}{0.95,0.95,0.95}

\newcommand{\todozlf}[2][]{\todo[color=yellow!80!black, #1]{LFZ: #2}}

\newcommand{\todonote}[1]{\todo[color=blue!50!white]{#1}}

\newcommand{\OursSpace}{BKLVA }
\newcommand{\OursNoSpace}{BKLVA}

\usepackage{amsmath,amsfonts,bm}

\def\eqref#1{equation~\ref{#1}}

\def\1{\bm{1}}

\DeclareMathAlphabet{\mathsfit}{\encodingdefault}{\sfdefault}{m}{sl}
\SetMathAlphabet{\mathsfit}{bold}{\encodingdefault}{\sfdefault}{bx}{n}

\def\gP{{\mathcal{P}}}

\def\sO{{\mathbb{O}}}

\usepackage{listings}
\definecolor{commentcolor}{rgb}{0.25,0.50,0.37}  %
\definecolor{keywordcolor}{rgb}{0.13,0.13,1}     %
\definecolor{stringcolor}{rgb}{0.63,0.13,0.94}   %
\definecolor{emphcolor}{rgb}{1,0.5,0}            %
\definecolor{mutedredorange}{rgb}{0.8, 0.4, 0.3}  %

\lstdefinelanguage{PDDL}{
  alsoletter={:,-,=},
  morekeywords={:action, :parameters, :domain, :precondition, :stream, :output, :certified, :effect, :effects, :ueffects, :conditions, :uconds, :axiom, :condition, :reward, not, and, possibly, or, imply, exists, forall, increase, when},
  morecomment=[l]{;},
  morestring=[b]",
}

\lstdefinestyle{PDDLStyle}{
  language=PDDL,
  basicstyle=\ttfamily\small, %
  breaklines=true,
  columns=fullflexible,
  keywordstyle=\color{keywordcolor}\bfseries,
  commentstyle=\color{commentcolor},
  stringstyle=\color{stringcolor},
  emph={maybe, not, and, or, imply, exists, forall, increase, when},
  emphstyle=\bfseries,
  showstringspaces=false, %
}

\lstdefinestyle{customlisp}{
  language=Lisp,
  basicstyle=\ttfamily\scriptsize,  %
  keywordstyle=\color{blue},
  commentstyle=\color{green!60!black},
  stringstyle=\color{purple},
  emphstyle=\bfseries,
  emph={action,parameters,precondition,effects,uconds,ueffects},
  morekeywords={and,maybe,not},
  numbers=none,
  literate=
    {:}{{\textcolor{blue}{:}}}1
    {-}{{\textcolor{blue}{-}}}1
    {?}{{\textcolor{red}{?}}}1
    {@}{{\textcolor{orange}{@}}}1,
}

\begin{document}

\title{Seeing is Believing: Belief-Space Planning\\ with Foundation Models as Uncertainty Estimators}

\author{
{Linfeng Zhao}{${^\P}$},
{Willie McClinton}{${^{*\S}}$},
{Aidan Curtis}{${^{*\S}}$},
{Nishanth Kumar}{${^\S}$},\\
{Tom Silver}{${^\ddagger}$},
{Leslie Pack Kaelbling}{${^\S}$},
{Lawson L.S. Wong}{${^\P}$}

\\
\vspace{0.2cm}
{${~^*}$}Equal Contribution,
{${~^\P}$}Northeastern University, 
{${~^\S}$}MIT,
{${~^\ddagger}$}Princeton University
}

\maketitle

\begin{abstract}
Generalizable robotic mobile manipulation in open-world environments poses significant challenges due to long horizons, complex goals, and partial observability. 
A promising approach to address these challenges involves planning with a library of parameterized skills, where a task planner sequences these skills to achieve goals specified in structured languages, such as logical expressions over symbolic facts. 
While vision-language models (VLMs) can be used to ground these expressions, they often assume full observability, leading to suboptimal behavior when the agent lacks sufficient information to evaluate facts with certainty.
This paper introduces a novel framework that leverages VLMs as a perception module to estimate uncertainty and facilitate symbolic grounding. 
Our approach constructs a symbolic belief representation and uses a belief-space planner to generate uncertainty-aware plans that incorporate strategic information gathering. 
This enables the agent to effectively reason about partial observability and property uncertainty.
We demonstrate our system on a range of challenging real-world tasks that require reasoning in partially observable environments. 
Simulated evaluations show that our approach outperforms both vanilla VLM-based end-to-end planning and VLM-based state estimation baselines, by planning for---and executing---strategic information gathering. 
This work highlights the potential of VLMs to construct belief-space symbolic scene representations, enabling downstream tasks such as uncertainty-aware planning.

\end{abstract}


\section{Introduction}

Task-level planning~\citep{tlp1989} is critical to achieving long-horizon mobile manipulation tasks in complex environments~\citep{homerobotovmmchallenge2023}.
However, standard planning methods typically assume complete knowledge of the environment, including property and quantity of objects, and rely on hand-crafted state estimators to acquire and track this information during execution.
Although these assumptions make planning easier, they limit the applicability to partially observable environments with ambiguous object properties and quantities.
Unlike fully observable environments where all objects and their properties are given, partially observable settings require robots to dynamically discover, perceive, and interact with objects while resolving uncertainties. 
This introduces a need for systems capable of handling not only object manipulation but also intertwined strategic information gathering to address unknowns in the environment.

\begin{figure}[t]
    \centering
    \small
    \includegraphics[width=0.45\textwidth]{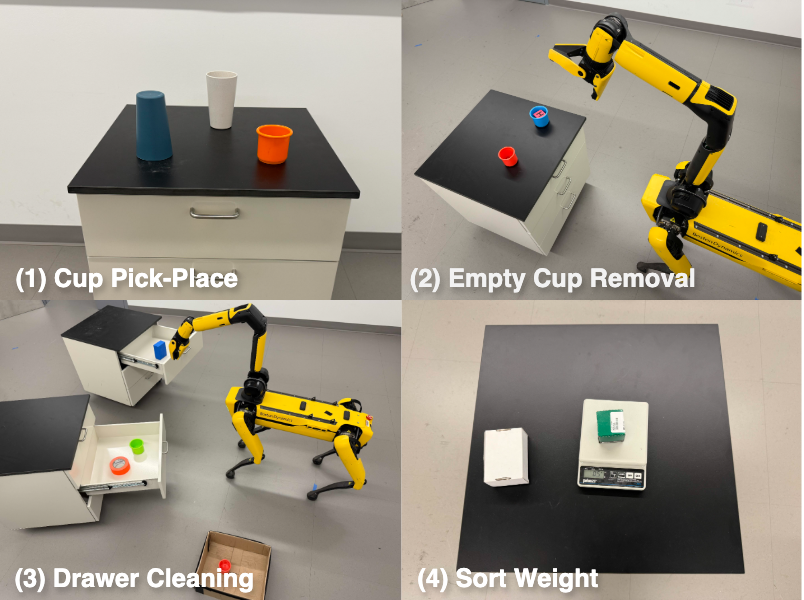}
    \caption{
    \small
    \textbf{Example tasks demonstrating various uncertainty levels.} 
    (1) Cup Pick-Place: a fully observable tabletop manipulation task with multiple cups.
    (2) Empty Cup Removal: requires inspecting cups from above to determine if they are empty before removal.
    (3) Drawer Cleaning: involves opening drawers to discover and remove objects inside.
    (4) Sort Weight: requires weighing sealed boxes on a scale to identify and dispose of empty ones.
    These tasks demonstrate increasing complexity in information gathering, from fully observable scenarios to those requiring strategic inspection and manipulation.}
    \label{fig:fig1-env-demo}
    \vspace{-10pt}
\end{figure}

We address the challenge of operating in partially observable environments characterized by three key aspects: (1) uncertain object properties that require strategic information gathering, (2) unknown number, type, and locations of object instances, and (3) goals specified as ungrounded natural language instructions.
In such domains, robots must plan not only manipulation actions, but also information-gathering actions. 
In Figure \ref{fig:fig1-env-demo}, we show some example tasks that require
handling unseen objects, properties, and quantities.

\begin{figure*}[!t]
    \centering
    \includegraphics[width=1.\textwidth]{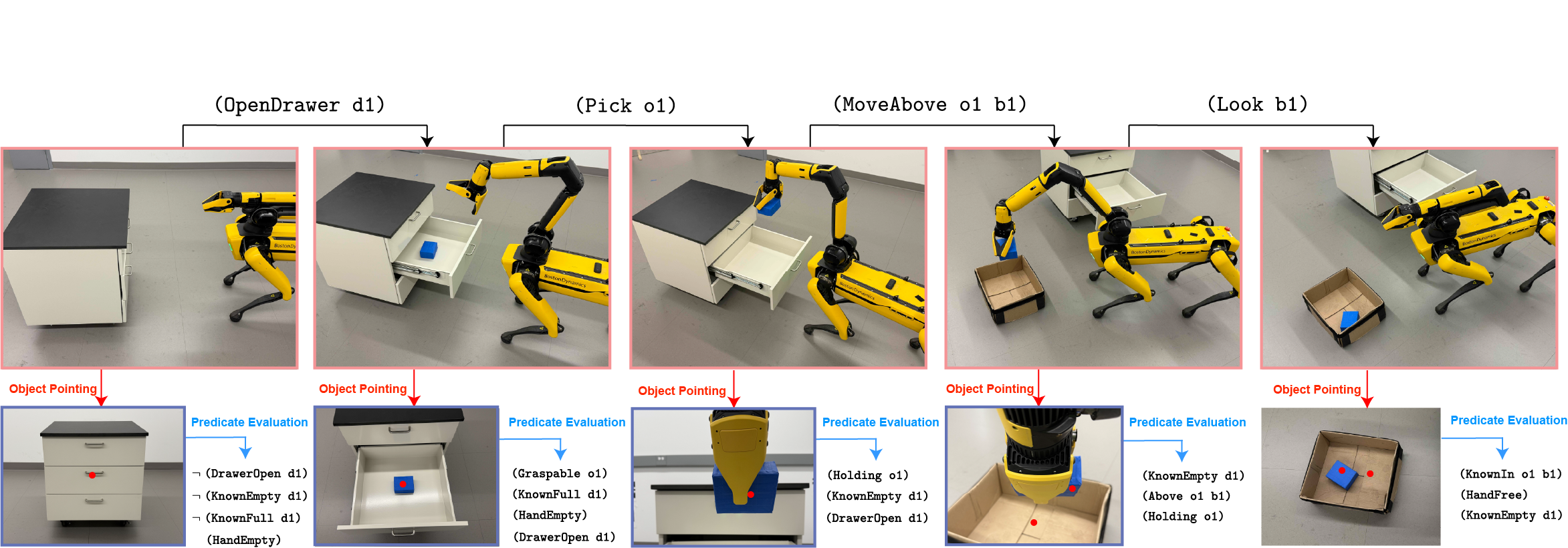}
    
    \caption{
    \small
    \textbf{Example plan.} A task to put any object in the drawer into a paper bin.
    Because the drawer is closed, the robot needs to maintain uncertainty of the environment and plan under uncertainty to achieve the belief goal: \texttt{KEmpty+(drawer)} and \texttt{Inside(block, box)}.
    The sequence shows: (1) initial reach to the closed drawer (without knowing if the drawer is empty or not), (2) opening the drawer to reveal a blue block inside and update belief, (3) grasping the block from the drawer, (4) moving the block over the paper bin, and (5) successfully placing the block into the bin. This demonstrates how the robot handles uncertainty through interleaved information gathering (opening drawer to check contents) and manipulation actions (grasping and placing the block).
    }
    \label{fig:example_plan}
    \vspace{-10pt}
\end{figure*}

Solving long-horizon problems in partially observable environments is particularly challenging. 
Two major strategies for handling partially observable environments include: (1) learning a policy directly over the entire observation history; and (2) aggregating information into a belief state for planning, such as in a belief-space task-and-motion planning (TAMP) system. Strategy 1 requires training history-conditioned policies from simulated or expert-provided data sources, which has exponentially increasing data requirements for increasingly long-horizon tasks. Strategy 2 requires planning in the belief space, which is currently achieved using hand-crafted belief-state estimators and belief-space transition models.

Previous approaches to planning under partial observability have relied on fully specified symbolic representations and predicates and used computationally expensive belief representations~\citep{kaelbling_integrated_2012,garrett_integrated_2020}, which require significant hand-engineering.
More recent work has explored foundation models like Vision-Language Models (VLMs) for planning. These approaches fall into Strategy 1, which typically requires a lot of training data to build a good vision-based state estimator and struggle with strategic information gathering and planning under uncertainty, especially for long-horizon real-robot tasks where data is very limited.

We present a light-weight strategy for belief-based planning in partially observable mobile-manipulation tasks. 
Our approach centers around representing the uncertainty of properties through three-valued predicates with known-true, known-false, and unknown values, and to systematically generate information gathering actions. 
This approach extends standard TAMP capabilities to handle partially observable scenarios while maintaining computational efficiency. 
We leverage VLMs both as flexible perception modules to evaluate arbitrary predicates on demand and to ground natural language goals into formal specifications.

We demonstrate our approach through %
experiments on mobile-manipulation tasks, in a synthetic environment with real images and on a physical Spot robot. These tasks require the agent to rearrange objects based on properties discovered during execution. Our results demonstrate that the system can effectively manage partial observability, dynamically adjust plans based on new information, and outperform existing methods in robustness and adaptability for simple manipulation tasks. In the synthetic tasks, our approach leverages strategic information gathering more efficiently than end-to-end counterparts. On a real robot, we showcase tasks where the robot strategically makes decisions under uncertainty and gathers information to solve long-horizon mobile-manipulation tasks. This highlights the potential of integrating VLMs with belief space planning for robust, easy-to-assemble, and adaptive robotic systems.

In summary, our work provides: (1) a general formalism for building a belief-space model for interleaved information gathering and mobile manipulation via a reduction to replanning in an augmented standard TAMP domain (such as~\citep{kumar_practice_2024}), (2) an integrated pipeline for a mobile-manipulation robot, that uses VLMs as a perception module and belief-space state estimator, and plans using the belief-space model to handle uncertainty in partially observable environments, and (3) demonstrations of strategic information gathering and decision-making under uncertainty in both synthetic tasks and real-world scenarios.

\section{Related Work}

Recent advances in pretrained foundation models, such as LLMs and VLMs, have improved high-level goal interpretation in robotics, but they struggle with systematic reasoning and uncertainty modeling. Belief-space planning and object-search methods address some of these challenges, but their integration with recent advances remains an active area of research.

\subsection{Foundation Models for Planning}

Recent advances in planning with LLMs and VLMs have enabled robots to perform approximate planning in partially observable environments~\cite{ahn2022icanisay, hazra2024saycanpayheuristicplanninglarge, curtis2024trustproc3ssolvinglonghorizon, liang2023codepolicieslanguagemodel}. However, recent evaluation efforts have shown the limitations of an approach without an explicit planning model in long-horizon and combinatorially difficult planning problems \citep{huang2022language, silver2022pddl}. These foundation models can be used in a ``zero shot'' way to directly generate plans from natural language goals and visual inputs, but they perform poorly on systematic reasoning and strategic decision-making \citep{huang2022inner}. Although they excel at understanding goals and scenes, they do not have effective mechanisms to maintain context in long action sequences and a long history of observations, reasoning about uncertainty, and handling tasks that require careful information gathering \citep{ahn2022can, valmeekam2023planning}. This will become particularly evident in our scenarios that require multistep reasoning about partially observable states or strategically choosing when and how to gather information about the environment.

\subsection{Belief-Space Planning}
Belief-space planning has been an effective method for handling environments with uncertainty due to partial observability~\citep{kaelbling_unifying_2012,kaelbling_integrated_2013,kaelbling2017pre,garrett2020online,curtis_partially_2024,curtis_long-horizon_2021,kaelbling2021specifyingachieving}, extending task and motion planning (TAMP) for handling long-horizon tasks \citep{garrett2021integrated}. These approaches model the robot's knowledge of the world as a belief state and plan to achieve goals described in terms of beliefs (e.g., "know that the lights are turned off") by combining actions that change the world state with actions that gather information.

Belief-based approaches generally have two modules: a state estimator that updates the belief based on the stream of observations coming into the system and a policy that maps the current belief into an action. One class of solutions constructs a complete policy before execution, which is capable of mapping any possible belief state to the optimal action.  This is generally computationally prohibitive and unnecessary, since only a tiny subset of possible beliefs will ever be reached. An alternative approach, which we follow, constructs partial plans online, and replans if, during execution, unanticipated observations are obtained~\citep{kaelbling_unifying_2012}.

Typically, these systems model the beliefs as probability distributions over possible world states; an alternative is to simply represent the belief as a set of possible environments. Such sets can be compactly represented in terms of a set of factors, such as ``door1 is open'' or ``there is an apple in the fridge'', each of which can be known to be true, known to be false, or unknown. This three-valued (ternary) logic has already been formalized \citep{ginsberg1988multivalued,kleene1952introduction,codd1979extending,zadeh1965fuzzy} and has been used in symbolic planning \citep{srivastava2008using}, giving an easier way to extend a deterministic transition model into one in belief space \citep{bonet2011}.

Current TAMP systems generally require a pre-specified domain of objects and properties, which makes them inapplicable to settings with open-ended goals and sets of objects.
Extending uncertainty-aware TAMP to include active information gathering and interaction with uncertainty about objects in partially observable settings is crucial for robots to operate in partially observable and long-horizon environments \citep{curtis_partially_2024}.
Sun et al. \citep{sun2024interactive} has used LLMs to generate plans, but relies on a hard-coded perception model to convert to the language description and does not explore task planning in the belief space in depth.

\subsection{Object Search}
One form of information gathering is Object search and exploration, which are a subclass of the mobile-manipulation problem space, and have been widely studied in the context of robotics and AI.
The methods used to solve these problems focus on locating objects in the environment, often using exploration strategies to discover objects that are not immediately visible~\citep{wong_manipulation-based_2013,wong_object-based_2015}. Although object search is essential for partially observable tasks, it is equally important to understand object properties and relationships to perform complex tasks effectively.
This work primarily studies the handling of object property-level uncertainty with belief-space planning instead of object search, enabling robots to reason about object properties and perform tasks that require a deeper understanding of the environment.
For example, to see if a cup is empty, the planner might have to remove the lid, place it on the table, and then look from above with its hand camera, taking multiple actions to perceive a single property.
This has been studied\todonote{Address Leslie's comment} in the literature with more classical systems without the use of modern perception systems \citep{sridharan2010planningtosee,srivastava_first-order_2014,nie2016searching}.

For goal parsing and scene understanding in planning, research has primarily focused on structured, fully observable environments \citep{silver2022pddl}.
These systems leverage object detection and LLMs to interpret high-level goals and generate symbolic representations of tasks. However, in partially observable environments, where both objects and their properties are often unknown, we must rely on \textit{Vision-Language Models (VLMs)} and active perception to dynamically discover and evaluate objects and their properties \citep{ahn2022can}.
It is essential to consider integrating active perception with belief-space planning for partially observable setups, which allows robots to perform planning to perceive: iteratively gathering information, refining their understanding, and performing tasks where objects and their properties cannot be assumed to be known in advance.

\section{Formulation: Modeling with Uncertainty}  %
\label{sec:formulation}

The problem of decision-making under uncertainty is typically modeled as a \textit{partially observable Markov decision process (POMDP)} \citep{kaelbling_planning_1998}, where an agent takes continuous actions and receives observations at each step, and the latent state space is continuous but with unknown dimension (state objects and their features). A standard approach is to convert the POMDP into a belief-space MDP, with a belief state representing a finite space of possible worlds with an infinite space of distributions over that finite space.

\subsection{Modeling the Environment with Uncertainty}
\label{subsec:environment-uncertainty}

In this work, we model the partially observable planning problem using an object-centric symbolic representation of the belief. We represent the system in terms of:
(1) a set of \textit{objects} $\sO$ that are conjectured  to exist;
(2) a set of \textit{predicates}, which are object-parameterized Boolean-valued relations over these objects and evaluated over the belief state, such as \texttt{Empty(cup)}, each of which can be \texttt{True}, \texttt{False}, or additionally \texttt{Unknown};
(3) a set of object-parameterized \textit{actions} that can be executed by the agent, each of which has preconditions and effects that modify the belief state; and
(4) a \textit{goal}, specified as a first-order logical expression over the relations that must be \texttt{True} in the terminal belief.

Each action has a \texttt{precondition} specifying when it is executable and an \texttt{effect} specifying how it modifies the belief state. For example, the action \texttt{Pick(cup)} is defined as follows in PDDL (Planning Domain Definition Language):
\begin{lstlisting}[style=PDDLStyle]
(:action Pick
 :parameters (?o - object)
 :precondition (and (HandEmpty) (CanGrasp ?o))
 :effects (and  ($\lnot$HandEmpty) (Holding ?o)))
\end{lstlisting}
In general, any action taken by the robot can both change the underlying world state and generate information that will reduce its uncertainty about the world state.  For simplicity in our model, we assume that these effects can be separated, so we model one class of actions as changing the world state deterministically but not yielding any additional information, and the other as not affecting the world state, but yielding information about some aspect of the state.   Furthermore, we assume (in the simplified planning model) that the observations are exactly correct and that there is no information loss due to the passage of time or the robot's actions.\footnote{Incorrect or unexpected observations are handled outside of belief-space planning, using replanning. This keeps the belief-space planning simple.}

\subsection{Belief Representation of Uncertainty}
\label{subsec:belief-representation}

In this work, we consider the uncertainty in the predicate values by representing it in the belief space through three-valued predicates.

\subsubsection{Three-Valued Predicates}

Following~\citet{bonet2011}, the three possibilities of a belief-space predicate can be represented by two binary predicates, through the use of known-predicates called $K$-fluents\footnote{Fluents are predicates that can be modified by the actions.} and actions that modify those $K$-fluents. 
For each predicate $P$, we define its known-true ($K_P$) and known-false ($K_{\neg P}$) counterparts, which track the agent's knowledge about the state of the world. 
If the value of $P$ is unknown, $K_P$ and $K_{\neg P}$ are both false, i.e., not known true and not known false. 
Actions intended to gather information on a predicate $P$ require $\lnot{K_P}\land\lnot{K_{\lnot{P}}}$ to hold in the precondition.

\begin{figure*}[t]
    \centering
    \includegraphics[width=0.95\textwidth]{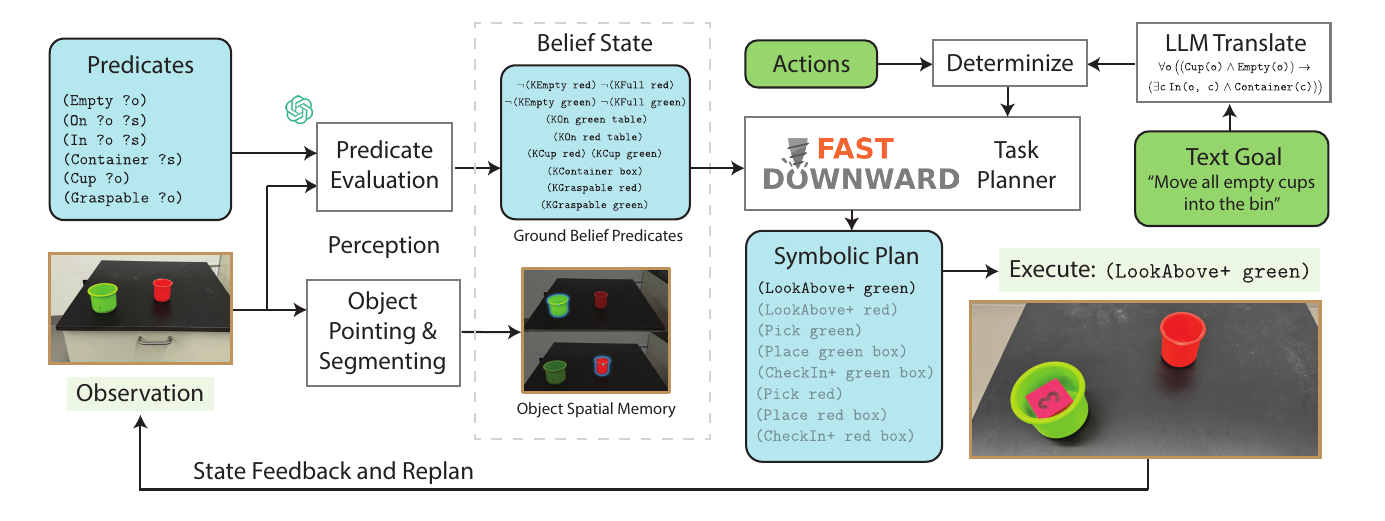}
    \caption{
        \small
        \textbf{Pipeline overview.} Our system integrates perception, belief-state update, and planning. 
        The example shows a task of moving empty cups to a bin, where the system must evaluate cup properties and plan appropriate manipulation actions.
        Before runtime, a text goal is first translated into symbolic specifications, which along with actions are determinized for the task planner.
        During a step of belief state update at runtime, given an observation (images and sensor inputs from a robot), the system performs two parallel processes for: (1) object pointing and segmenting to maintain a spatial memory of objects, and (2) predicate evaluation to ground belief predicates (e.g., \texttt{Empty}, \texttt{On}).
        The planner generates a symbolic plan based on the symbolic belief state, and the first action is executed to generate a new observation.
        The belief state is updated based on the new observation, and the process repeats until the goal is satisfied.
    }
    \label{fig:pipeline_overview}
    \vspace{-10pt}
\end{figure*}

\subsubsection{Information-Gathering Actions}

An information-gathering action can be modeled using a PDDL operator description~\citep{fox2003pddl2}, except that there are no specified truth values for effects, such as \texttt{ObserveEmptiness(?cup)}.
Since the effect of an information-gathering action can either be $K_P$ or $K_{\lnot{P}}$, the agent's transition model become nondeterministic. 
This nondeterministic planning problem can be translated into a deterministic planning by splitting a single nondeterministic action resulting in either $K_P$ (e.g., \texttt{KEmpty+}) or $K_{\lnot{P}}$ (e.g., \texttt{KEmpty-}) into two independent deterministic actions $A_{+}$, $A_{-}$ (e.g., \texttt{ObserveEmptiness+}, \texttt{ObserveEmptiness-}) resulting in $K_P$ and $K_{\lnot{P}}$, respectively. 
For example, an action that results in a belief-changing observation may look as follows after an optimistic determinization (which allows the planning agent to ``choose'' the outcome), where the effect is a conjunction of \texttt{KEmpty+} and \texttt{KEmpty-}:
 \todonote{Address Leslie's comment}

\begin{lstlisting}[style=PDDLStyle]
(:action ObserveEmptiness+
 :parameters (?o - object ?s - surface)
 :precondition (and (On ?o ?s) (HandEmpty) 
                        ($\lnot$KEmpty+ ?o) ($\lnot$KEmpty- ?o))
 :effects (and (KEmpty+ ?o))
\end{lstlisting}

\begin{lstlisting}[style=PDDLStyle]
(:action ObserveEmptiness-
 :parameters (?o - object ?s - surface)
 :precondition (and (On ?o ?s) (HandEmpty) 
                        ($\lnot$KEmpty+ ?o) ($\lnot$KEmpty- ?o))
 :effects (and (KEmpty- ?o))
\end{lstlisting}

This is an optimistic determinization because it allows the planning agent to ``choose'' the outcome by selecting the appropriate action.
However, the actual outcome may not be expected, and the agent can observe the environment and replan.
Under some assumptions it has been shown that, if replanning is performed after each execution step, the solution to the determinized problem is equivalent to the solution to its nondeterminisic counterpart~\cite{bonet2011}.

\subsection{Representation of Belief State}

There are several parts of the belief state:
(1) the set of objects conjectured to exist,
(2) the set of symbolic predicates evaluated over the objects given current knowledge,
(3) physical information, such as the locations of the objects and the robot's state (e.g., its gripper state).
The first two parts are abstracted into a symbolic belief state, similar to PDDL, \todonote{Address Leslie's comment} and the third part is to represent additional physical information.

During execution time, the robot plans and acts repeatedly until the goal is satisfied under its belief.
A goal is a first-order logical expression over the predicates that must be \texttt{True} in the terminal belief.
As the robot makes observations, new objects may be added to the belief, and more relations become definitively \texttt{True} or \texttt{False}. For simplicity, we assume that the world dynamics are deterministic and perception is perfectly accurate, so that once an object is added or a relation is observed to be \texttt{True} or \texttt{False}, it remains so unless explicitly changed by the robot.
While our approach does not model information loss over time, it provides a practical balance between expressiveness and computational tractability.

\textit{Incidental Object Discovery:}
Our approach achieves incidental object discovery through a property-based mechanism rather than direct object search. We define predicates that describe the state of locations where objects might be present. For example, the predicate \texttt{EmptyContainer(?drawer)} represents whether a drawer contains any object: if it is \texttt{True}, the drawer is believed to be empty; if it is \texttt{False}, the drawer is believed to contain some object; if it is \texttt{Unknown}, the robot needs to perform an information-gathering action to determine its status.
By utilizing this three-valued logic system for location predicates, when a location is discovered to be non-empty (i.e., \texttt{EmptyContainer(?drawer)} becomes \texttt{False}), newly discovered objects may be added to the set of conjectured objects $\sO$.
This strategy enables incidental object discovery when interacting with containers or locations, allowing the robot to make informed decisions when new objects are revealed. However, it does not support active search for objects in unknown locations or conjecture about objects that have not been observed.

\begin{algorithm}[t]
    \caption{Belief-Space Planning and Execution (\S\ref{sec:execution-pipeline})}
    \label{alg:belief_space_planning}
    \begin{algorithmic}[1]
    \Require Initial belief state $b_0$ with initial objects $\mathbb{O}_0$, text goal $g_\text{text}$, predicates $\mathcal{P}$, actions $\mathcal{A}$
    \State $g \gets \texttt{Translate}(g_\text{text}, \mathcal{P}, \sO_0)$ \Comment{Translation (\S\ref{subsec:execution-overview})}
    \State $b \gets b_0$
    \While{$\lnot \texttt{Satisfied}(g, b)$}
        \State $p \gets \texttt{Plan}(b, g, \mathcal{A})$ \Comment{Generate plan (\S\ref{subsec:planning-execution})}
        \State \textbf{if} $p = \text{None}$ \textbf{then} \Return False \Comment{Goal is infeasible.}
        \For{$a$ \textbf{in} $p$}
            \State $o \gets \texttt{Execute}(a)$ \Comment{Get observation (\S\ref{subsec:vlm-perception})}
            \State $b \gets \texttt{BeliefUpdate}(b, a, o)$ \Comment{Update (\S\ref{subsec:belief-update})}
            \If{$\lnot \texttt{ExpectedEffects}(a, b)$}
                \State \textbf{break} \Comment{Replan with updated belief (\S\ref{subsec:planning-execution})}
            \EndIf
        \EndFor
    \EndWhile
    \State \Return \texttt{True} \Comment{Goal achieved}
    \end{algorithmic}
\end{algorithm}

\section{Runtime: Planning under Uncertainty}
\label{sec:execution-pipeline}

In this section, we detail the process of the pipeline for planning under uncertainty, where we term BKLVA: \textit{\textbf{B}elief-space planning with \textbf{K}-fluents, \textbf{L}LM-based goal-grounding, \textbf{V}LM-based perception and estimation, and information-gathering \textbf{A}ctions}.
The execution pipeline takes as input:
(1) a belief-space domain with observation operators,
(2) a natural language goal $g_\text{text}$,
(3) pretrained foundation models for perception, and
(4) an initial set of known objects and their properties.

\subsection{Pipeline Overview: Observe-Update-Plan-Execute}
\label{subsec:execution-overview}

The execution pipeline begins by translating the natural language goal $g_\text{text}$ into a lifted first-order logical expression $G$ using the predicates $\gP$ defined in our belief-space domain. Importantly, these goals can be specified over both known and yet-to-be-discovered objects. For example, ``move any objects in the drawer to the bin'' would translate to $\forall x.\texttt{InBin}(x)$, where $x$ could match objects discovered in the drawer during execution. This lifted representation allows the system to handle partially observable scenarios where the complete set of objects is not initially known.
An alternative solution is to translate to object-grounded logical expression (e.g.,\texttt{InBin(cup1)}), which needs to be updated when new objects are found.

To begin execution, we process initial observations $o^\prime$ from the robot's starting position or a brief exploration routine. These observations are processed by our VLM perception system (see \S\ref{subsec:vlm-perception}) to:
(1) identify objects of interest $\sO$ and their geometric properties (e.g., locations, spatial extents) in a global coordinate frame, and
(2) evaluate predicates $\gP$ over these objects to determine whether each is known-true ($K_P$), known-false ($K_{\neg P}$), or unknown ($\neg K_P \land \neg K_{\neg P}$).
This constructs our initial belief state $b$.

The execution then proceeds in an observe-plan-execute cycle:

\textit{(1) Perception and Belief Update}
The VLM processes observations $o^\prime$ to detect objects $\sO^\prime$ and their geometric information. New objects are merged with existing ones \todonote{Explain in appendix} ($\sO \gets \sO \cup \sO^\prime$) using a data association based on their identifying characteristics (location and description). The belief state is updated by reevaluating predicates, where only the predicates which can be confidently determined by the VLM (known-true or known-false) are updated. For example, after observing inside a drawer, we might update \texttt{KEmpty+(drawer1)}.

\textit{(2) Planning}
Given the current belief state $b$, the symbolic planner (Fast Downward \cite{helmert2006fast}) generates a plan $p$ using goal $G$, action descriptions $A$, and object set $\sO$. The planner works with a determinized domain where observation actions are split into optimistic outcomes. For example, \texttt{LookInDrawer} is split into \texttt{LookInDrawer+} (assuming empty) and \texttt{LookInDrawer-} (assuming not empty). If no valid plan exists, the goal is deemed infeasible.

\textit{(3) Execution}
The first action $p[0]$ is executed (e.g., \texttt{OpenDrawer(drawer1)}). If the action's outcome differs from the planner's optimistic assumption (e.g., drawer not empty when assumed empty), we trigger replanning.

This cycle continues until either the belief state satisfies the goal ($b \models G$) or the goal is determined to be infeasible. After each action, if the outcome differed from expectations, we replan from the current belief state to adapt to the new situation.

\subsection{Perception via Vision-Language Models}
\label{subsec:vlm-perception}

Our perception system utilizes Vision-Language Models (VLMs) to process visual observations and extract both object detections and predicate evaluations. The perception pipeline consists of two main components:

\subsubsection{Object Detection and Localization}
For object detection, we utilize MOLMO~\citep{deitke2024molmopixmoopenweights}, a pre-trained model with a pointing feature that enables object localization. Given an observation $\bar{o}$ and a set of textual prompts of objects, MOLMO identifies objects of interest $\sO$, providing their pixel locations in the image. These detections are then processed using the Segment Anything Model (SAM)~\citep{kirillov2023segment} to generate segmentation masks, which combined with depth information, yield spatial extents $S$ and locations $L$ in a global coordinate frame (maintained by an underlying SLAM-based odometry system).

\subsubsection{Predicate Evaluation}
For each detected object, we query a VLM (e.g., GPT-4o~\citep{openai_gpt-4_2023}) to evaluate the truth values of predicates over these objects. For example, given an image of a drawer, we might ask ``Is the drawer empty?'' The VLM acts as a belief-space grounding classifier, determining whether each predicate is known-true ($K_P$), known-false ($K_{\neg P}$), or remains unknown ($\neg K_P \land \neg K_{\neg P}$). This evaluation is done in a batched manner for all predicates on all images observations at the current time step for efficiency, with a single query evaluating multiple predicates simultaneously.
For example, a Spot robot has 6 cameras, and we ground all predicates using the objects in the system to the 
\todozlf{discuss why this isn't costly in compute and time; each skill should only need to call VLM twice at the beginning (for checking precondition) and termination (for checked expected effect); can say a skill takes > 1 VLM call; explain in appendix}
\todozlf{add more details}

\subsection{Belief State Update}
\label{subsec:belief-update}

The belief-state update process integrates new observations with the existing belief state, maintaining consistency in object tracking and predicate evaluations:

\subsubsection{Object List Update}
When new observations $o^\prime$ yield object detections, we merge them with existing objects using data association based on their positions in the global coordinate frame. For example, if we detect ``cup4'' near a previously known ``drawer1'', we add it to our object set while maintaining spatial relationships. This process ensures consistent object tracking while avoiding duplication.

\subsubsection{Predicate Value Update}
For all objects visible in the current observation, we update the truth values of predicates based on the VLM's evaluations. For example, after observing inside a drawer, we might update $\texttt{EmptyContainer(drawer1)-}$.
The belief update mechanism follows a monotonic knowledge acquisition principle: unknown predicates ($\neg K_P \land \neg K_{\neg P}$) can transition to known states ($K_P$ or $K_{\neg P}$), but known predicates never revert to unknown states. Additionally, we maintain a quasi-static assumption where predicate values remain unchanged unless explicitly modified by new observations or actions.

\subsection{Execution and Replanning}
\label{subsec:planning-execution}

\subsubsection{Execution}
Plan execution involves both high-level symbolic action selection and additionally parameter generation of the parameterized skills.
For high-level actions, we execute the first action $p[0]$ from the current plan.
\todozlf{polish the writing on the low level part based on Tom's comments}
For low level, the object parameters (e.g., \texttt{PlaceOn(cup1, bin0)}) are converted to continuous parameters using either learned models~\citep{kumar_practice_2024} or VLM-based parameter suggestion (see Appendix for details). The execution module maintains geometric consistency and handles physical constraints through a motion planning layer.

\subsubsection{Replanning}
Replanning is triggered in the following scenarios:
(1) when an observation action yields an unexpected outcome (e.g., finding a drawer non-empty when assumed empty), or
(2) when new objects are incidentally discovered.
When replanning is triggered, we update the belief state with the new information and generate a new plan from the current state.
\todozlf{update this}

\section{Experiments}

\begin{figure*}[!t]
    \centering
    \includegraphics[width=1.\linewidth]{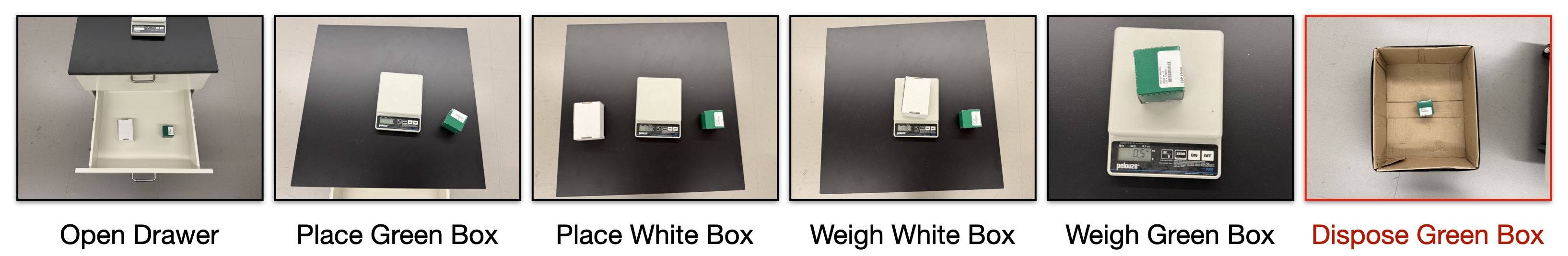}
    \caption{
        \small
        \emph{Sort Weight}. An example task in our synthetic environment with real images. The agent needs to open the drawer and retrieve sealed boxes to weigh them. The boxes cannot be opened by the agent but can only be measured indrectly by a scale. The goal is to find empty boxes and remove it to a bin, and a few notable states are shown in the figure. The optimal path takes 14 steps.
    }
    \label{fig:placeholder-experiment}
\end{figure*}

\begin{table*}[t]
    \centering
    \resizebox{\textwidth}{!}{%
    \begin{tabular}{r|cccccc}
    \toprule
    \multirow{2}{*}{\diagbox[width=16em]{\textbf{Methods}}{\textbf{Tasks}}} & \multicolumn{2}{c}{\textbf{Cup Pick-Place}} & \multicolumn{2}{c}{\textbf{Drawer Cleaning}} & \multicolumn{2}{c}{\textbf{Sort Weight}} \\
    \cmidrule(lr){2-3} \cmidrule(lr){4-5} \cmidrule(lr){6-7}
                                    & Success       & SPL           & Success & SPL            & Success & SPL \\
    \midrule
    \textbf{Random}                 & $0\%$         & $0.00\pm0.00$ & $0\%$   & $0.00\pm0.00$  & $0\%$   & $0.00\pm0.00$ \\
    \textbf{VLM End-to-End}         & $30\%$        & $0.15\pm0.24$ & $0\%$   & $0.00\pm0.00$  & $0\%$   & $0.00\pm0.00$ \\
    \textbf{VLM (Captioning) + LLM} & \textbf{100\%}       & $0.49\pm0.07$ & $10\%$  & $0.04\pm0.11$  & $0\%$   & $0.00\pm0.00$ \\
    \textbf{VLM (Labeling) + LLM}   & $90\%$        & $0.69\pm0.28$ & $0\%$   & $0.00\pm0.00$  & $0\%$   & $0.00\pm0.00$ \\
    \textbf{\OursNoSpace{} (Ours)}  & \textbf{100\%} & \textbf{1.00 $\pm$ 0.00} & \textbf{80\%} & \textbf{0.32 $\pm$ 0.16} & \textbf{70\%} & \textbf{0.46 $\pm$ 0.32} \\
    \bottomrule
    \end{tabular}
    }
    \caption{
    \small
    Performance comparison across synthetic tasks. Success indicates success rate (\%) and SPL indicates average success rate weighted by (normalized inverse) path length (between 0 to 1, with 1 being optimal).
    }
    \label{tab:sim_results}
    \end{table*}

Through our experiments, we aim to answer several key questions:
(i) Does our structured approach (using VLMs for belief state estimation for symbolic planning) effectively handle uncertainty?
(ii) How efficient is our belief-space planning strategy compared to alternatives?
(iii) Can this pipeline extend to a real-world robot scenario with real perception and control?

\textbf{Environments.}
We evaluate the approaches on (1) a synthetic environment with real images for mobile manipulation, and (2) a Spot real-robot mobile-manipulation environment.
A synthetic task is defined within a fully-observable transition graph that accepts symbolic actions and returns both images and other non-visual predicates (e.g,. whether the agent is holding an object or reachable to an object), though the images may not reveal the entire system state (for example, a drawer's contents remain hidden from view until it is opened). Consequently, similar to a real-robot scenario, the agent maintains a partially observable belief model to plan effectively. We build synthetic domains using real-world images (taken by phones or robots), where each node in the environment's transition graph represents a distinct viewpoint (represented as images and other non-visual predicates). When the agent selects an action, the environment presents images of the resulting state, allowing the agent to update its belief, which initially may contain uncertain or incomplete knowledge of object properties. This symbolic environment with real images (1) systematically evaluates symbolic belief-space planning that integrates perception and task reasoning, (2) reduces the complexity of physical control, and (3) manages randomness when comparing against multiple baselines. These features enable us to explore long-horizon belief-space planning tasks.
\todo{add spot env a bit?}

We evaluate on the following tasks:
\begin{itemize}[leftmargin=*]
    \item \textit{Cup Pick-Place (Synthetic).} This is a fully-observable task that tests the agent's ability to rearrange some cups on a table into a box. A set of information-gathering actions is provided to verify if an agent understands whether uncertainty is present and needed to be reduced.

    \item \textit{Drawer Cleaning (Synthetic).} In this task, one or more drawers of cabinets may contain various objects. Initially, the robot does not know which drawers hold items. It must open each drawer to observe its contents, dynamically update its belief regarding newly found objects, and then remove those objects to a box. An illustration with one drawer and one block is shown in Figure~\ref{fig:example_plan}. The synthetic task features 2 objects and 1 drawer.

    \item \textit{Sort Weight (Synthetic).} The agent needs to remove closed boxes by using a scale to measure the weight of the boxes. The boxes are hidden inside cabinets, so the agent needs to find the cabinets and measure their weight. After finding the empty boxes, the agent needs to remove them to a bin.

    \item \textit{Empty Cup Removal (Robot).}
    A Spot robot is used to execute this task, where three cups are placed on two tables, while the robot does not know if cups contain contents or are empty. From a normal front-facing view, the robot cannot distinguish whether a cup is empty or not. It must navigate closer, take a camera perspective from above, and inspect the contents. Once the robot identifies an empty cup, it removes it to a bin. We include this setup as a real-robot demo of the system, integrated with real-world object detection, segmentation, and belief-state update.
    See Figure~\ref{fig:pipeline_overview} and the supplementary video.

\end{itemize}

\textbf{Approaches.}
We evaluate the performance of our approach in comparison to several baselines across simulation and real-robot environments. The environments vary in terms of observability and task complexity, including both fully observable and partially observable settings, as well as short-horizon and long-horizon tasks. 
The approaches include:
\todonote{add pointer to appendix}
\begin{itemize}[leftmargin=*]
    \item \textit{Random} planner that selects object-parameterized skills and valid object parameters uniformly at random.
    \item \textit{VLM End-to-end Planning} uses VLMs for end-to-end planning without explicit handling of uncertainty. It has been explored in tabletop manipulation tasks and web agent literature.
    \item \textit{VLM State Captioning + LLM Planning} uses VLM to generate a text caption of the history of observations (with images and other non-visual predicates). An LLM then outputs a sequence of actions using the caption.
    \item \textit{VLM Predicate Labeling + LLM Planning} uses VLM to perceive predicate values and LLM for planning, but does not handle uncertainty explicitly either.
    \item \textit{\OursSpace(Ours)}: An approach that uses belief-space operators integrated with VLM-based belief-state estimation to plan to gather information in order to achieve the goal.
\end{itemize}
For VLM-based planner or VLM-based state captioning, we provide the history of visual observations to them for handling partial observability.
Replanning is used and needed when the expected outcome is not achieved, particularly in partially observable environments where belief-space operators are determinized to an optimistic outcome.
See the appendix for more details.
\todonote{remove variants for now}

\textbf{Experimental Setup.}
For each synthetic task, we run 10 random seeds and report the average results, controlling for variance in the perception system, foundation model calls, planning, low-level control, and other factors.
For all LLM- and VLM-based approaches, we use \texttt{GPT-4o} with zero-shot prompting, with available objects, operators parameterized by objects, operators' preconditions and effects, current state (LLM-based), and observation (VLM-based) or history (partially observable variants) provided in the context window.
\todonote{some more important setup}
We use the following metrics to evaluate the approaches for performance and efficiency in handling uncertainty in completing tasks: (1) task success rate, (2) average number of symbolic actions task plan length required for successful runs of a task weighted by the success rate, also referred as ``SPL'' (Success rate weighted by Path Length) in visual navigation.
The agent succeeds when it reaches the goal state and fails when it reaches the max number of steps\todonote{elaborate}, transitions to an illegal state (e.g., pick up full cup), or gets stuck in a state.
For \textit{real-robot} experiments, we use the Spot robot and only demonstrate with our approach.
We use the same set of symbolic actions (operators) as the synthetic tasks, including the information-gathering and manipulation actions. Each action is additionally associated with a skill that executes on the robot. More details are provided in the appendix.

\textbf{Results and Discussion.}
We first evaluate whether our structured approach, which integrates VLM-based predicate estimation with symbolic planning, effectively handles uncertainty (\textbf{Q1}). As shown in Table~\ref{tab:sim_results}, our method consistently outperforms baselines on partially observable tasks like \emph{Drawer Cleaning} and \emph{Sort Weight}. Unlike end-to-end VLM methods or caption-based state representations, our system selectively gathers information only when needed (e.g., opening drawers or weighing boxes if relevant), thereby avoiding redundant actions. 

We also compare the efficiency of our belief-space planner against alternatives on gathering information (\textbf{Q2}).
We notice a few patterns that baselines do not handle well and result in lower success rates and longer SPL: (1) it relies on commonsense knowledge to plan, which may take extra steps based on its commonsense; (2) it does not capture some subtle diffences in state, causing failure or inefficiency on the task; (3) it does not necessarily understand the history and may retry the same actions at same or similar states that occurred before; (4) it does not output correct format of the state.
As an example, the \emph{Cup Pick-Place} task provides information-gathering actions while all information is provided, all baselines tend to take extra steps and result in lower SPL, because it does not understand the actions can be directly executed without additional information needed.
A hypothetical approach that separates information gathering and manipulation stages by exhaustively removing uncertainty (e.g., opening all drawers first) would be prohibitively time-consuming in large or intricate environments.
By focusing on only those uncertainty necessary to fulfill the task goals, our method achieves higher success rates with fewer actions.

Finally, we test whether the proposed pipeline extends to real-robot scenarios on the \emph{Empty Cup Removal} task (\textbf{Q3}), with a video provided in the supplementary material. A Spot robot inspects three cups placed on different tables to determine which are empty before disposing of them. The same VLM-based predicate evaluation prompts are used as in synthetic tasks on images. 
Our demonstration shows that the pipeline of VLM-based belief state estimation and symbolic planning in BKLVA transfers well to real-world perception and control despite additional noise and uncertainty, as seen in the accompanying video. Thus, the real-robot trials confirm that our framework remains effective in physical settings, suggesting strong potential for more complex mobile-manipulation tasks beyond laboratory conditions.

Overall, these results validate each of our three research questions: (1) explicit predicate-based reasoning and belief maintenance of BKLVA enables robust handling of uncertainty, (2) belief-space planning improves efficiency compared to naive or exhaustive approaches, and (3) the BKLVA pipeline scales to a real robot with minimal modification (besides additional real-world perception and control), demonstrating its viability for real-world partial observability.

\section{Limitations and Conclusion}

In this work, we have assumed an \textit{a priori}, human-defined modeling stage for operator-level transitions. While this assumption significantly simplifies system integration, it may limit adaptability when the environment or task changes. In principle, future approaches could partly automate or learn these operators, or extract from pretrained foundation models, alleviating the need for manual specification.

Additionally, our approach relies on a perception system based on pretrained vision-language models whose performance bounds the overall success of our planning framework. The symbolic or LLM-based planners and language-based goal specifications ultimately depend on detected object symbols. Although we partially address this by using a VLM to propose newly discovered objects, full object search remains a substantial challenge, requiring long-horizon exploration, active perception, and robust low-level control. Integrating these elements into a more comprehensive pipeline is an important avenue for future work.

Finally, we have not deeply explored the intricacies of low-level skill integration, such as motion planning under physical constraints or uncertainty in control. Existing integrated TAMP methods provide some building blocks, but bridging high-level belief-space planning with real-time skill execution demands carefully vetted assumptions about robot dynamics and hardware. In future work, we plan to leverage these TAMP frameworks more thoroughly, enabling a tighter coupling between symbolic reasoning and physical execution.

\section{Conclusion}

In this paper, we presented a novel approach to belief-space robot planning that effectively addresses the challenges of partial observability and uncertainty in partially observable environments on mobile-manipulation robots. By integrating belief-space planning with information-gathering actions and leveraging vision-language models (VLMs) as belief-state estimators, our method demonstrated superior performance in handling long-horizon tasks in diverse, partially observable settings. The ability to reason about information gathering using ternary belief-space predicates enabled the robot to systematically reduce uncertainty, leading to higher task-success rates and improved efficiency compared to baseline approaches.

Our results highlight the importance of combining high-level belief-space reasoning with robust perception systems in robotic planning frameworks. Unlike baselines that lacked explicit uncertainty handling or efficient planning for information gathering, our approach demonstrated the advantage of integrating symbolic planning with modern perception systems, particularly VLMs. By automating the process of belief-state estimation with VLMs, we made belief-space planning more feasible in real-world scenarios, addressing the complexity of manually designing predicates for dynamic, unstructured environments.

This work serves as a foundation for future research in belief-space planning. While we focused on leveraging VLMs for belief-state estimation, future extensions could explore learning belief-space operators, improving sampling strategies, and integrating end-to-end learning to bridge the gap between planning and execution. These advancements could further enhance the practicality of belief-space planning and enable more adaptive and capable robotic systems in partially observed settings.

\bibliographystyle{unsrtnat}
\bibliography{ref_linfeng_zotero,references}

\clearpage

\section{Extended Setup Information}
\label{sec:appendix:setup}

\subsection{Environment-Agent Interface for Spot}
We employ the Boston Dynamics Spot robot, which is equipped with six RGBD cameras. One camera is mounted on the manipulator arm, providing a close-up view for fine manipulation tasks. The remaining five are body cameras (two front-facing, one left-facing, one right-facing, and one rear-facing), enabling a 360° view for object detection, tracking, and situational awareness.

The Python SDK for Spot is used to send velocity and motion commands to the robot, receive feedback on joint states and robot pose, and retrieve RGB and depth images from all six cameras. By leveraging this SDK, we can synchronize robot actions with sensor data retrieval, ensuring that each execution step is accompanied by timely, high-quality visual feedback.

\subsection{Parameterized Skill Execution}
Our approach separates planning decisions from low-level control. A classical symbolic planner outlines sequences of actions (like `pick` or `place`), but the details of each action—such as determining grasp points or collision-free paths—come from parameterized robot skills. 

For instance, when the robot needs to pick an object, we first verify that its gripper is free. Then, we figure out where to grab the object, how much force to apply, and what collision checks are needed. These skill parameters are fed into low-level controllers that oversee individual movements and sensor checks. 

We find that splitting tasks in this manner keeps the planner's high-level reasoning clean and avoids entangling it with control-specific intricacies~\citep{silver2022learning,kumar_practice_2024}. In future work, we plan to detail the underlying motion planners and control loops used for these skill executions, as well as refine thresholds like approach velocity or force limits.

\subsection{Simulated Environment Setup}

We develop a real-image interactive symbolic environment (RISE) to replicate tasks without running on the real robot. This environment is partially observable and uses pre-captured images---one image per state---to provide synthetic RGBD inputs for our planner and visual pipeline. Each state is annotated with which objects are currently visible, whether the robot's gripper is free or holding an object, and other relevant properties. By linking these states with ``transitions'' that correspond to high-level actions (e.g., \texttt{PickObjectFromTop} or \texttt{MoveToHandViewObject}), we can define which actions can succeed or fail from a given state.

The primary data structure keeps track of:
\begin{enumerate}
    \item Unique State IDs %
    \item Associated camera images (for simulating ``views'' of the environment)
    \item A record of which objects appear in that view and which, if any, the robot is currently holding
    \item A set of transitions connecting one state to another, each labeled with the operator name and whether the action succeeds or fails
\end{enumerate}

For instance, if the planner issues a \texttt{PickObjectFromTop} action in the ``initial'' state, the mock environment checks whether there is a corresponding transition from ``initial'' under \texttt{PickObjectFromTop}. If so, it moves us to the new state (e.g., ``holding\_block'') and returns updated mock sensor data. Because transitions are explicitly defined, we can also encode negative outcomes---like failing to pick an object---by directing the environment to a ``pick\_failed'' state.

To better test tasks involving uncertainty, we add ``belief operators.'' These can require or modify belief-related predicates (e.g., \texttt{Unknown\_ContainerEmpty}) instead of purely physical ones. For example, an \texttt{ObserveContainerContent} operator can transition the system from an unknown content state to a known one, based on an imagined ``camera check.''

Because this environment structure stays close to the real system's domain model, it uncovers logical flaws before deployment. We also minimize discrepancies by matching coordinate frames and rough sensor noise levels wherever possible. In doing so, we ensure that end-to-end tests in this mock environment---involving both symbolic planning and VLM-based perception---translate smoothly onto the physical Spot robot.

\section{Methods and Baselines Details}
\label{sec:appendix:methods}

\begin{table*}[!t]
    \centering
    \begin{tabular}{r|cccc|cc}  %
        \toprule
        \rowcolor{white} \textbf{Components} & \multicolumn{4}{c|}{\textbf{Current-State Uncertainty: Perception}} & \multicolumn{2}{c}{\textbf{Future-State U.: Planning}} \\
        \rowcolor{white} & \textbf{Goal Grounding} & \textbf{O. Existance} & \textbf{O. Property} & \textbf{Belief Classifier} & \textbf{Belief-Space} & \textbf{Long-Horizon} \\
        \midrule
        BHPN \citep{kaelbling_integrated_2013} & \cellcolor{orange!20}? & \cellcolor{green!20}\checkmark & \cellcolor{green!20}\checkmark & \cellcolor{orange!20}? & \cellcolor{green!20}\checkmark & \cellcolor{green!20}\checkmark \\
        EES \citep{kumar_practice_2024} & \cellcolor{red!20}\texttimes & \cellcolor{red!20}\texttimes & \cellcolor{red!20}\texttimes & \cellcolor{red!20}\texttimes & \cellcolor{red!20}\texttimes & \cellcolor{green!20}\checkmark \\
        \midrule
        VLM End-to-End & \cellcolor{green!20}\checkmark & \cellcolor{orange!20}? & \cellcolor{red!20}\texttimes & \cellcolor{red!20}\texttimes & \cellcolor{orange!20}? & \cellcolor{red!20}\texttimes \\
        VLM (Text) + LLM Plan & \cellcolor{green!20}\checkmark & \cellcolor{orange!20}? & \cellcolor{orange!20}? & \cellcolor{orange!20}? & \cellcolor{orange!20}? & \cellcolor{red!20}\texttimes \\
        \textbf{VLM (Predicate)} + LLM Plan & \cellcolor{green!20}\checkmark & \cellcolor{orange!20}? & \cellcolor{green!20}\checkmark & \cellcolor{green!20}\checkmark & \cellcolor{orange!20}? & \cellcolor{red!20}\texttimes \\
        \midrule
        \textbf{Ours} & \cellcolor{green!20}\checkmark & \cellcolor{orange!20}? & \cellcolor{green!20}\checkmark & \cellcolor{green!20}\checkmark & \cellcolor{green!20}\checkmark & \cellcolor{green!20}\checkmark \\
        \bottomrule
    \end{tabular}
    \caption{Comparison of different approaches based on uncertainty handling and planning capabilities. O. stands for object.}
    \label{tab:approaches_comparison}
    \vspace{-10pt}
\end{table*}

\subsection{Comparison of Approaches}
To assess the performance of our system, we compare it with several baselines. End-to-end policy learning attempts to map raw pixels directly to actions without explicit symbolic inference; while flexible, it often struggles with task generalization and replanning. Classical planners without visual reasoning rely entirely on predetermined symbols and have difficulty adjusting when environmental assumptions become invalid. Reactive execution systems use local triggers without a global notion of a task-level plan, leading to limited adaptability.

In Table~\ref{tab:approaches_comparison}, we summarize several approaches in terms of how they address ``current-state uncertainty'' (object existence, object property classification, etc.) and ``future-state uncertainty'' (belief-space planning, long-horizon tasks). 
For instance, BHPN~\citep{kaelbling_integrated_2013} models partial observability and can handle unknown object states through belief-space planning, but it relies primarily on hand-coded perception and does not natively ground goals from language. 
EES~\citep{kumar_practice_2024} leverages a symbolic planner for complicated tasks but does not explicitly deal with object existence or property uncertainty and requires an enumerated domain.

VLM End-to-End approaches (e.g.,~\citep{ahn2022can}) can parse language goals but generally do not perform systematic belief updating or fully handle uncertainty about concealed or unknown objects. 
Similarly, VLM(Text)+LLM Plan can handle language-based goals and partially reason about objects but typically lacks explicit belief updates and relies on ad-hoc textual prompts to track state. 
A stronger variant, VLM(Predicate)+LLM Plan, uses predicate-level queries on images but lacks robust planning under unknown states or objects.

Our method (bottom in the table) combines all these features: it (1) accepts language-derived goals and newly discovered objects, (2) systematically updates beliefs about object existence or properties using VLM-based classifiers, and (3) runs a full belief-space planner for strategic information gathering and long-horizon tasks. This unified approach gives it comprehensive coverage of both current-state uncertainties (like unknown container contents) and future-state uncertainties (planning how and when to gather information).

\section{Additional Implementation Details}
\label{sec:appendix:implementation}

\subsection{Visual Predicates Evaluated by VLM}
As part of our system, we define a set of predicates that are evaluated using the Visual Language Model (VLM). Each predicate has a name, a list of argument types (written as ?movable, ?container, etc.), and a specific prompt text. The prompt guides the VLM in determining whether the predicate is true based on a given image or scene description. Below is a bullet list of the main VLM-based predicates we use:

\begin{itemize}
    \item \textbf{On}(\texttt{?movable}, \texttt{?base}) \\
    \textbf{Prompt:}  
    ``This predicate describes when a movable object is on a flat surface. It conflicts with the object being Inside a container. Please check the image and confirm the object is on the surface. If it's truly on top (e.g., on a table or floor) and not inside something else, answer yes. Otherwise, answer no.''

    \item \textbf{Inside}(\texttt{?movable}, \texttt{?container}) \\
    \textbf{Prompt:}  
    ``Use this predicate when an object is inside a container and not just resting on a surface. If you see the object's shape overlapping the container's interior, answer yes. If it's merely on top or partially overlapping, answer no.''

    \item \textbf{Blocking}(\texttt{?base}, \texttt{?base}) \\
    \textbf{Prompt:}  
    ``Check if one object is blocking the robot from accessing or viewing another. If so, answer yes; if not, answer no.''

    \item \textbf{NotBlocked}(\texttt{?base}) \\
    \textbf{Prompt:}  
    ``Confirm that no object is blocking the given object. If you see no obstruction, answer yes. Otherwise, answer no.''

    \item \textbf{NotInsideAnyContainer}(\texttt{?movable}) \\
    \textbf{Prompt:}  
    ``This predicate is true if the object is not inside any container. If it is inside something, answer no.''

    \item \textbf{InHandViewFromTop}(\texttt{?robot}, \texttt{?movable}) \\
    \textbf{Prompt:}  
    ``Answer yes if the robot's camera is positioned above the movable object to see into it (e.g., looking inside a cup). If unsure or the view is angled, answer no.''

    \item \textbf{Unknown\_ContainerEmpty}(\texttt{?container}), \textbf{Known\_ContainerEmpty}(\texttt{?container}), \textbf{BelieveTrue\_ContainerEmpty}(\texttt{?container}), \textbf{BelieveFalse\_ContainerEmpty}(\texttt{?container}) \\
    \textbf{Prompts:}  
    ``[Answer: yes/no only]
    (1) \emph{Unknown\_ContainerEmpty}: You do not have enough information to decide if the container is empty.  
    (2) \emph{Known\_ContainerEmpty}: You are confident whether it is empty or not.  
    (3) \emph{BelieveTrue\_ContainerEmpty}: You believe the container is empty (e.g., you see only a single color inside).  
    (4) \emph{BelieveFalse\_ContainerEmpty}: You believe the container has contents (multiple colors or visible items).''

    \item \textbf{Unknown\_Inside}(\texttt{?movable}, \texttt{?container}), \textbf{Known\_Inside}(\texttt{?movable}, \texttt{?container}), \textbf{BelieveTrue\_Inside}(\texttt{?movable}, \texttt{?container}), \textbf{BelieveFalse\_Inside}(\texttt{?movable}, \texttt{?container}) \\
    \textbf{Prompts:}  
    ``[Answer: yes/no only]
    (1) \emph{Unknown\_Inside}: You are uncertain whether the object is inside the container.  
    (2) \emph{Known\_Inside}: You can confidently tell if it is inside or not.  
    (3) \emph{BelieveTrue\_Inside}: You believe the object is fully inside the container.  
    (4) \emph{BelieveFalse\_Inside}: You believe the object is not inside (e.g., it is on top or separate).''

\end{itemize}

We also group certain predicates into belief categories or container-related predicates, but the fundamental structure remains the same: each predicate is evaluated by the VLM in response to a carefully written prompt that captures how to determine truth from the image.

\subsection{World-State and Belief-Space Operators}
\label{sec:pddl-operators}

Below, we present a representative set of operators in a PDDL-style pseudocode. Each operator includes parameters, preconditions, and effects, with line breaks and indentation for readability.

Note that these operators are shared for the synthetic RISE tasks and for the real-robot tasks.

\begin{lstlisting}[style=PDDLStyle]
(:action MoveToReachObject
 :parameters (?robot - Robot ?object - BaseObject)
 :precondition (and
   (NotBlocked ?object)
   (NotHolding ?robot ?object)
 )
 :effect (and
   (Reachable ?robot ?object)
))

(:action MoveToHandViewObject
 :parameters (?robot - Robot ?object - Movable)
 :precondition (and
   (NotBlocked ?object)
   (HandEmpty ?robot)
 )
 :effect (and
   (InHandView ?robot ?object)
))

(:action MoveToBodyViewObject
 :parameters (?robot - Robot ?object - Movable)
 :precondition (and
   (NotBlocked ?object)
   (NotHolding ?robot ?object)
 )
 :effect (and
   (InView ?robot ?object)
))

(:action PickObjectFromTop
 :parameters
   (?robot - Robot ?object - Movable ?surface - Immovable)
 :precondition (and
   (On ?object ?surface)
   (HandEmpty ?robot)
   (InHandView ?robot ?object)
   (NotInsideAnyContainer ?object)
   (IsPlaceable ?object)
   (HasFlatTopSurface ?surface)
 )
 :effect (and
   (Holding ?robot ?object)
   (not (On ?object ?surface))
   (not (HandEmpty ?robot))
   (not (InHandView ?robot ?object))
   (not (NotHolding ?robot ?object))
))

(:action PlaceObjectOnTop
 :parameters
   (?robot - Robot ?held - Movable ?surface - Immovable)
 :precondition (and
   (Holding ?robot ?held)
   (Reachable ?robot ?surface)
   (NEq ?held ?surface)
   (IsPlaceable ?held)
   (HasFlatTopSurface ?surface)
   (FitsInXY ?held ?surface)
 )
 :effect (and
   (On ?held ?surface)
   (HandEmpty ?robot)
   (NotHolding ?robot ?held)
   (not (Holding ?robot ?held))
))

(:action MoveToHandViewObjectFromTop
 :parameters (?robot - Robot ?object - Movable)
 :precondition (and
   (NotBlocked ?object)
   (HandEmpty ?robot)
 )
 :effect (and
   (InHandViewFromTop ?robot ?object)
   (InHandView ?robot ?object)  ; derived from the top view
))

(:action ObserveCupContentFindNotEmpty
 :parameters (?robot - Robot ?cup - Container ?surface - Immovable)
 :precondition (and
   (On ?cup ?surface)
   (InHandViewFromTop ?robot ?cup)
   (HandEmpty ?robot)
   (NotHolding ?robot ?cup)
   (Unknown_ContainerEmpty ?cup)
 )
 :effect (and
   (Known_ContainerEmpty ?cup)
   (BelieveFalse_ContainerEmpty ?cup)
   (not (Unknown_ContainerEmpty ?cup))
))

(:action ObserveCupContentFindEmpty
 :parameters (?robot - Robot ?cup - Container ?surface - Immovable)
 :precondition (and
   (On ?cup ?surface)
   (InHandViewFromTop ?robot ?cup)
   (HandEmpty ?robot)
   (NotHolding ?robot ?cup)
   (Unknown_ContainerEmpty ?cup)
 )
 :effect (and
   (Known_ContainerEmpty ?cup)
   (BelieveTrue_ContainerEmpty ?cup)
   (not (Unknown_ContainerEmpty ?cup))
))

(:action ObserveDrawerEmpty
 :parameters (?robot - Robot ?container - Container)
 :precondition (and
   (Unknown_ContainerEmpty ?container)
   (DrawerOpen ?container)
   (Reachable ?robot ?container)
 )
 :effect (and
   (Known_ContainerEmpty ?container)
   (BelieveTrue_ContainerEmpty ?container)
   (not (Unknown_ContainerEmpty ?container))
))

(:action ObserveDrawerNotEmpty
 :parameters (?robot - Robot ?container - Container)
 :precondition (and
   (Unknown_ContainerEmpty ?container)
   (DrawerOpen ?container)
   (Reachable ?robot ?container)
 )
 :effect (and
   (Known_ContainerEmpty ?container)
   (BelieveFalse_ContainerEmpty ?container)
   (not (Unknown_ContainerEmpty ?container))
))
\end{lstlisting}

\subsection{Prompts to Pretrained Models}

Below are example prompts used to query our vision-language and language models:

\noindent
\textbf{VLM Predicate Evaluation Prompt}

\begin{lstlisting}[style=PDDLStyle]
Your goal is to answer questions related to object relationships in the 
given image(s) from the cameras of a Spot robot. Each question is independent
while all questions rely on the same set of Spot images at a certain moment.

We will use the following predicate-style descriptions to ask questions:
    Inside(object1, container)
    Blocking(object1, object2)
    On(object, surface)

Some predicates may include 'KnownAsTrue' or 'KnownAsFalse'.
You should respond 'Yes' or 'No' but never 'Unknown'.
If you don't know the answer for 'KnownAsTrue' or 'KnownAsFalse' predicates,
say 'No'.

Here are VLM predicates we have, note that they are defined
over typed variables. Example: (<predicate-name> <obj1-variable>:<obj1-type> ...)

VLM Predicates (separated by line or newline character):
{vlm_predicates}

Examples (separated by line or newline character):
Do these predicates hold in the following images?
1. Inside(apple:object, bowl:container)
2. On(apple:object, table:surface)
3. Blocking(apple:object, orange:object)
4. Blocking(apple:object, apple:object)
5. On(apple:object, apple:object)
6. On(apple:object, bowl:container)
7. EmptyKnownTrue(bowl:container)
8. EmptyKnownFalse(bowl:container)
9. Inside(bowl:container, bowl:container)

Answer with explanation and Yes/No for each question. Keep each explanation
and answer in a single line, with no empty lines between responses:
1. I can see the apple is clearly contained within the bowl's interior. [Yes]
2. The apple appears to be floating above the table, not making contact. [No]
3. The apple is positioned directly in front of the orange, preventing access. [Yes]
4. ... [No]
5. ... [No]
6. ... [Yes]
7. ... [Yes]
8. ... [No]
9. ... [No]

Actual questions (separated by line or newline character):
Do these predicates hold in the following images?
{question}

Answer with explanation and Yes/No for each question. Keep each explanation and
answer in a single line, with no empty lines between responses:
\end{lstlisting}

\noindent
\textbf{LLM Planner Prompt}

\begin{lstlisting}[style=PDDLStyle]
You are highly skilled in robotic task planning, breaking down intricate and long-term tasks into distinct primitive actions.
Consider the following skills a robotic agent can perform. Note that each of these skills takes the form of a `ParameterizedOption` and may have both discrete arguments (indicated by the `types` field, referring to objects of particular types),
as well as continuous arguments (indicated by `params_space` field, which is formatted as `Box([<param1_lower_bound>, <param2_lower_bound>, ...], [<param1_upper_bound>, <param2_upper_bound>, ...], (<number_of_params>,), <datatype_of_all_params>)`).

{options}

Preconditions indicate the properties of the scene that must be true for you to execute an action. The effects are what will happen to the scene when you execute the actions.
You are only allowed to use the provided skills. It's essential to stick to the format of these basic skills. When creating a plan, replace
the arguments of each skill with specific items or continuous parameters. You can first describe the provided scene and what it indicates about the provided
task objects to help you come up with a plan.

Here is a list of objects present in this scene for this task, along with their type (formatted as <object_name>:<type_name>):
{typed_objects}

And here are the available types (formatted in PDDL style as `<type_name1> <type_name2>... - <parent_type_name>`). You can infer a hierarchy of types via this:
{type_hierarchy}

Here is the current state of the scene:
{state_str}

Finally, here is an expression corresponding to the current task goal that must be achieved:
{goal_str}

Here is the history of actions executed so far (if any):
{action_history}

Please return a plan that achieves the provided goal from the current state.
Please provide your output in the following format:
1. First write "Explanation of scene + your reasoning" followed by your explanation
2. Then write "Plan:" on a new line
3. Then write each action on a new line in EXACTLY this format (no numbers, no code blocks, no extra formatting):
skill_name(object1:type1, object2:type2)[param1, param2]

For example:
Explanation of scene + your reasoning
This is a simple pick and place task where we need to...

Plan:
MoveToObject(robot:robot, cup:movable_object)[]
PickObject(robot:robot, cup:movable_object)[]
MoveToLocation(robot:robot, table:surface)[]
PlaceObject(robot:robot, cup:movable_object, table:surface)[]
OpenDrawer(robot:robot, drawer:container)[]

Do not include any numbers, bullet points, code blocks, or other formatting. Just write the plan exactly as shown above.
...
\end{lstlisting}

\noindent
\textbf{VLM Planner Prompt}

\begin{lstlisting}[style=PDDLStyle]
You are highly skilled in robotic task planning, breaking down intricate and long-term tasks into distinct primitive actions.
Consider the following skills a robotic agent can perform. Note that each of these skills takes the form of a `ParameterizedOption` and may have both discrete arguments (indicated by the `types` field, referring to objects of particular types),
as well as continuous arguments (indicated by `params_space` field, which is formatted as `Box([<param1_lower_bound>, <param2_lower_bound>, ...], [<param1_upper_bound>, <param2_upper_bound>, ...], (<number_of_params>,), <datatype_of_all_params>)`).

{options}

You are only allowed to use the provided skills. It's essential to stick to the format of these basic skills. When creating a plan, replace
the arguments of each skill with specific items or continuous parameters. You can first describe the provided scene and what it indicates about the provided
task objects to help you come up with a plan.

Here is a list of objects present in this scene for this task, along with their type (formatted as <object_name>: <type_name>):
{typed_objects}

And here are the available types (formatted in PDDL style as `<type_name1> <type_name2>... - <parent_type_name>`). You can infer a hierarchy of types via this:
{type_hierarchy}

Finally, here is an expression corresponding to the current task goal that must be achieved:
{goal_str}

Here is the history of actions executed so far (if any):
{action_history}

Please return a plan that achieves the provided goal from an initial state depicted by the image(s) below.
IMPORTANT: You must follow this EXACT format (including exact spacing and newlines):

Explanation of scene + your reasoning
<your explanation here>

Plan:
<action1>
<action2>
...

Each action must be in this exact format with no extra spaces or formatting:
skill_name(object1:type1, object2:type2)[param1, param2]

Example output:
Explanation of scene + your reasoning
The robot needs to pick up a cup from the table and place it on the shelf.

Plan:
MoveToObject(robot:robot, cup:movable_object)[]
PickObject(robot:robot, cup:movable_object)[]
MoveToLocation(robot:robot, shelf:surface)[]
PlaceObject(robot:robot, cup:movable_object, shelf:surface)[]

CRITICAL: 
- Do not add any numbers, bullet points, asterisks, or code blocks
- Do not add any extra newlines between actions
- Write "Plan:" exactly like that, with the colon and one newline after
- Each action must be on its own line with no extra formatting    
\end{lstlisting}

\end{document}